\pdfoutput=1

\documentclass[11pt]{article}

\usepackage{ACL2023}

\usepackage{times}
\usepackage{latexsym}

\usepackage[T1]{fontenc}

\usepackage[utf8]{inputenc}

\usepackage{microtype}

\usepackage{inconsolata}

\usepackage{graphicx} 
\usepackage{amsmath,amssymb} %
\usepackage{bbm}
\usepackage{multirow}
\usepackage{subcaption}
\usepackage{caption}
\usepackage{microtype}
\usepackage{graphicx}
\usepackage{times}
\usepackage{latexsym}
\usepackage{amsmath}
\usepackage{amsmath,algorithm,float,tabularx,amsfonts}
\usepackage{algpseudocode}
\makeatletter
\newcommand{\multiline}[1]{%
    \begin{tabularx}{\dimexpr0.9\linewidth-\ALG@thistlm}[t]{@{}X@{}}
        #1
    \end{tabularx}
}
\usepackage[T1]{fontenc}
\usepackage{booktabs}
\usepackage[utf8]{inputenc}

\usepackage{microtype}
\usepackage{graphicx}
\usepackage{inconsolata}
\title{Large Scale Generative Multimodal Attribute Extraction for \\E-commerce Attributes}

\author{Anant Khandelwal \Thanks{This work was done while author was in International Machine Learning team.} \\
  Ads Trust \\
  Amazon \\
  \texttt{anantkha@amazon.com} \\\And
  Happy Mittal \\
  International Machine Learning \\
  Amazon \\
  \texttt{mithappy@amazon.com} \\
  \AND
  Shreyas Sunil Kulkarni\\
  International Machine Learning \\
  Amazon \\
  \texttt{kulkshre@amazon.com} \\
  \And
  Deepak Gupta\\
  International Machine Learning \\
  Amazon \\
  \texttt{dgupt@amazon.com} \\
}

\begin{document}
\maketitle
\begin{abstract}
    E-commerce websites (e.g. Amazon) have a plethora of structured and unstructured information (text and images) present on the product pages. Sellers often either don't 
label or mislabel values of the attributes (e.g. color, size etc.) for their products. Automatically identifying these attribute values from an 
eCommerce product page that contains both text and images is a challenging task, especially when the attribute value is not explicitly mentioned in the catalog. In this 
paper, we present a scalable solution for this problem where we pose attribute extraction problem as a question-answering task, which we solve 
using \textbf{MXT}, consisting of three key components: (i) \textbf{M}AG (Multimodal Adaptation Gate), (ii) \textbf{X}ception network, and (iii) \textbf{T}5 encoder-decoder. Our system consists of a 
generative model that \emph{generates} attribute-values for a given product by using both textual and visual characteristics (e.g. images) of the product. We show that our system is capable of handling zero-shot attribute prediction (when attribute value is not seen in training data) and value-absent prediction (when attribute value is not mentioned in the text) which are missing in traditional classification-based and NER-based models respectively.
We have trained our models using distant supervision, removing dependency on human labeling, thus making them practical for real-world applications. With this framework, we are able to train a single model for 
1000s of (product-type, attribute) pairs, thus reducing the overhead of training and maintaining separate models. Extensive experiments on two real world datasets show that our framework improves the absolute recall@90P by 10.16\% and 6.9\% from the existing state of the art models. In a popular e-commerce store, we have deployed our models for 1000s of (product-type, attribute) pairs.
\end{abstract}
\section{Introduction}
\begin{figure}[h]
  \centering
    \includegraphics[scale=0.3]{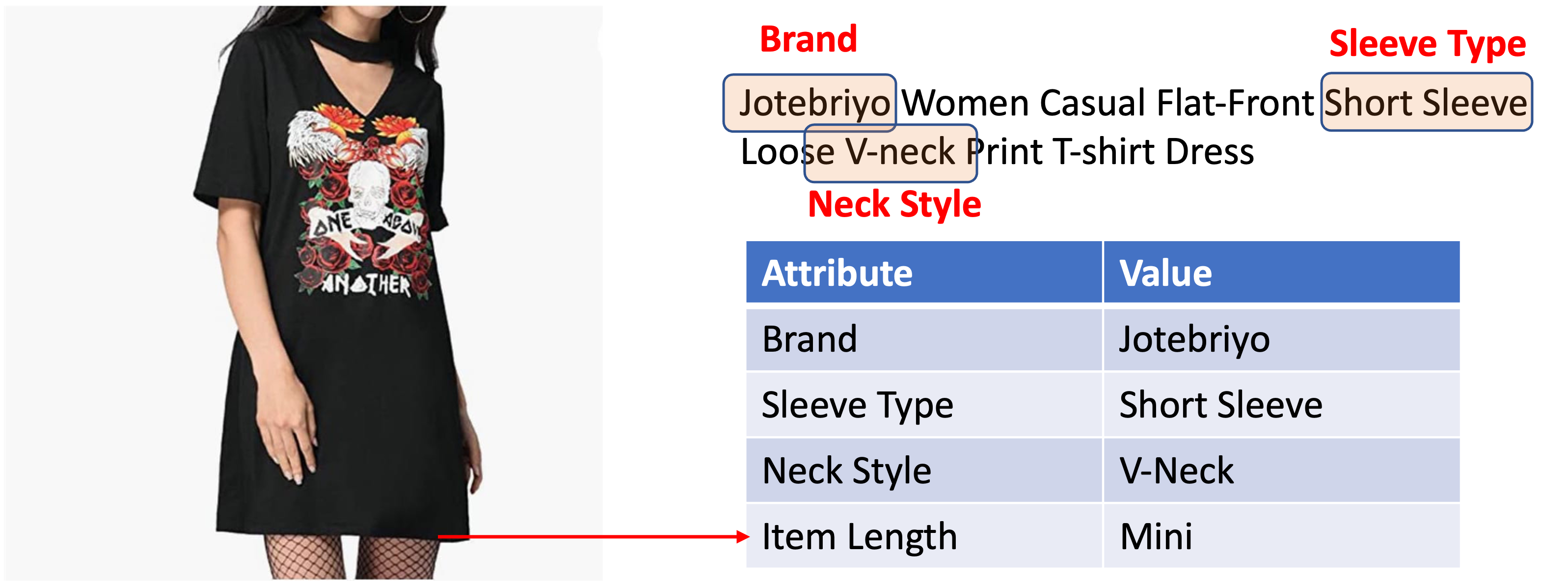}
    \caption{Illustration of attribute extraction problem}
  \label{fig:intro-fig}
\end{figure}

E-commerce websites (e.g. Amazon, Alibaba) have a very wide catalog of products. Seller provided catalog of these products contain both textual information and product images. Apart from 
this unstructured information, they also provide structured information about the products such as color, material, size, etc. This information can be represented in terms of attribute-value pairs (see figure~\ref{fig:intro-fig}). In this paper, we will use the terms attribute and attribute-name interchangeably. The value of attribute will be referred as \emph{attribute-value}. 
However, while listing the products, sellers rarely specify all attribute values or mistakenly fill incorrect values. These attribute
values may or may not be present in the unstructured textual product information. Extracting/inferring the missing attribute values from the unstructured textual product
information (and images) can improve the catalog quality, thereby improving the customer experience (again, refer figure~\ref{fig:intro-fig} for an example of attribute extraction).

\textbf{PT-attribute:} A PT-attribute is defined as a pair of $(\text{product-type},\text{attribute})$, where product-type (or PT) is a broad category of products (e.g. "shoes", "dress", "laptops" etc.) and attribute is an attribute-name (e.g. "color", "size" etc.). Typically, attribute-extraction is done at the granularity of PT-attribute (e.g. "extract the value of \emph{color} attribute of \emph{shoe}").

A good attribute extraction system has following desirable properties: (1) \textbf{Scalability:} A single model should handle multiple PT-attributes so that there is no need to train a separate model for every PT-attribute combination, (2) \textbf{Multi-modality:} Model should be able to extract attributes from multiple modalities like text, image, video etc., (3) \textbf{Zero-shot inference:} Model should be able to extract attribute values that were not seen in the training data, and (4) \textbf{Value-absent inference:} Model should extract attribute values that are not explicitly mentioned in the text on the product page (but can be inferred from image or some other reasoning). 

\textbf{Related Work: }Extensive research has been done to build attribute extraction models, which can be categorized as \textit{extractive}, \textit{predictive}, or \textit{generative}. Extractive models pose 
this problem as a Named Entity Recognition (NER) problem~\cite{Zheng2018OpenTagOA}. Some of the recent work in this space include
LATEX-numeric~\cite{Mehta2021LATEXNumericLA}, 
and MQMRC~\cite{Shrimal2022NERMQMRCFN} 
. However, these models don't do value-absent inference. Moreover,
these are text based models and do not use product images. Predictive models are the classifier models that take text 
(and image) as input and predict the attribute values. CMA-CLIP~\cite{Liu2021CMACLIPCA} is a recent multi-modal predictive framework for predicting attribute values. However, these models can't do zero-shot inference as the prediction comes from the predefined classes only.
Generative models pose this problem as an answer generation task given a question and context. Here, the question is the attribute name, and context
is the product data (text and image), and the answer is the attribute value. For example, Roy et. al.~\cite{roy-etal-2021-attribute} presented a generative framework to
generate attribute values using product's text data. PAM~\cite{Lin2021PAMUP} introduced a multi-modal generative framework, however their model 
requires (i) Training encoder and decoder from scratch, (ii) Manually modifying the vocabulary of outputs (attribute-values) for different product-types.

In this paper, we present \textbf{MXT}, a multimodal generative framework to solve the attribute extraction problem, that consists of three key components: (i) \textbf{M}AG (Multimodal Adaptation Gate)~\cite{rahman-etal-2020-integrating}: a fusion framework to combine textual and visual embeddings, that enables generating image-aware textual embeddings, (ii) \textbf{X}ception network~\cite{chollet2017xception}: an image encoder that generates attribute-aware visual embeddings, and (iii) \textbf{T}5 encoder-decoder~\cite{Raffel2020ExploringTL}. The models trained by our generative framework are scalable as a single model is trained on 
multiple PT-attributes, thus reducing the overhead of training and maintaining separate models. We remove the disadvantages of PAM model by (i) finetuning a strong pre-trained language model 
(T5~\cite{Raffel2020ExploringTL}) and thus leveraging its text generation ability, (ii) providing product-type in the input itself so that output distribution
is automatically conditioned on the PT. Moreover, our trained model satisfies all of the 4 desirable properties that were mentioned previously. 

\begin{figure*}[h]
  \centering
    \includegraphics[width=\textwidth]{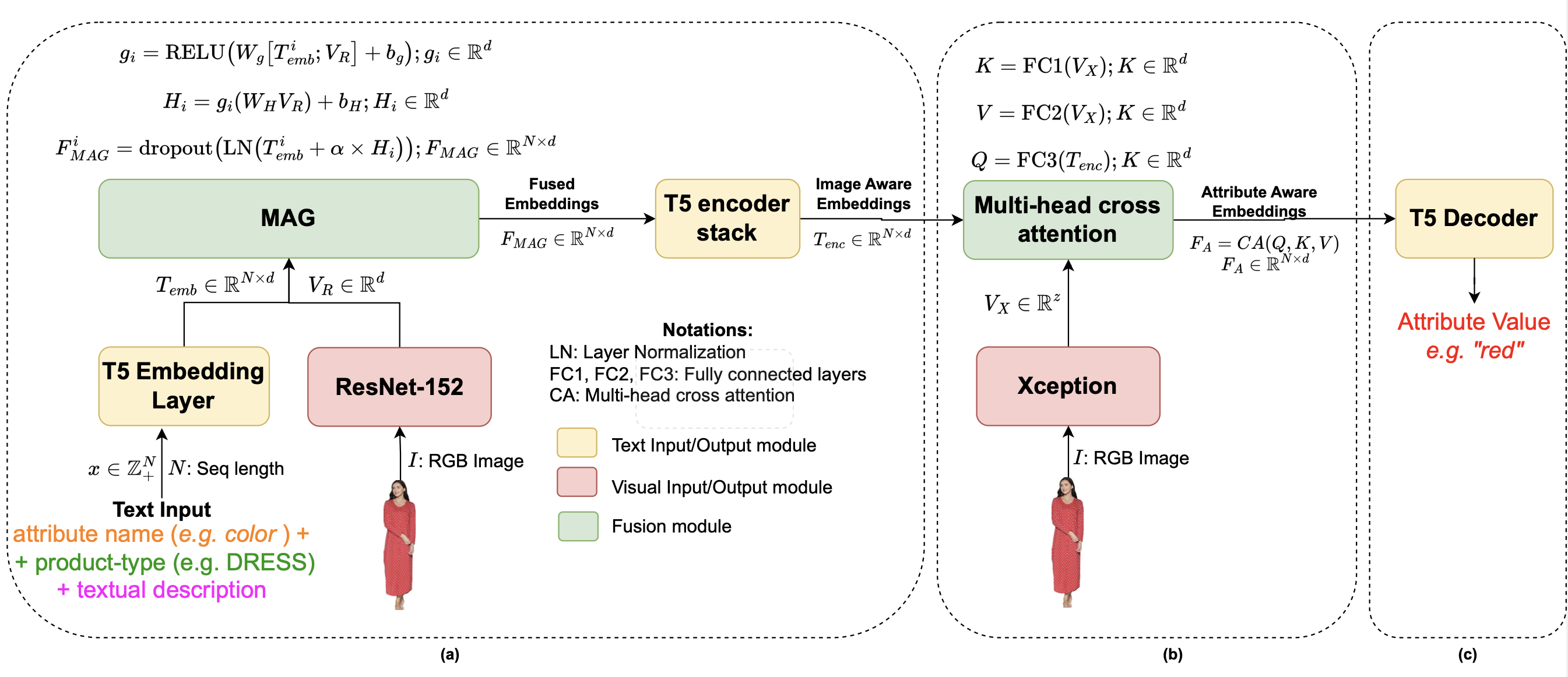}
    \caption{Architecture of MXT. (a) Generates image-aware text embeddings by fusing 
    image embeddings (obtained from ResNet-152) and text embeddings of the input text (concatenation of \emph{attribute name}, \emph{product type}, 
 and textual description of the product), (b) Image-aware text embeddings are then attended with region specific 
    visual embeddings obtained from separable convolution of Xception Network, which in turn passes only the attribute specific embeddings 
    to the decoder (c) Fused embeddings are passed through T5 decoder to generate 
    attribute value.}
  \label{fig:mmpag_arch}
\end{figure*}

Our system formulates the attribute extraction problem as a question-answering problem, where (a) question is the
attribute name (e.g. "color"), (b) textual context comprises of a concatenation of product-type (e.g. "shirt"), and textual description of the product, (c) visual context comprises product image, and (d) answer is the attribute value for the attribute
specified in the question. Our model architecture consists of (i) a T5 encoder to encode the question and textual context, (ii) encoding visual context into product specific embeddings
 through a pre-trained ResNet-152 model~\cite{He2016DeepRL} and fusing them with T5's textual embeddings using a multimodal adaptation gate 
 (MAG)~\cite{Rahman2020IntegratingMI}, (iii) encoding visual context into attribute (e.g. "sleeves", "collar" etc.) specific embeddings through 
 Xception model~\cite{chollet2017xception} and fusing
them with previously fused embeddings through a dot product attention layer~\cite{Yu2021VisionGG}, and finally (iv) generating the attribute values through T5 decoder. The detailed architecture of our system is shown
 in figure~\ref{fig:mmpag_arch}. 
 
In section~\ref{sec:mxt}, we explain our proposed model MXT. In section~\ref{sec:experiments}, we compare our model's performance 
with NER-Based MQMRC~\cite{shrimal-etal-2022-ner} along with a popular multi-modal model CMA-CLIP~\cite{Liu2021CMACLIPCA} and show that on same precision, we outperform them (on recall) for a majority of the attributes. We also show an ablation study justifying the proposal of different components in MXT. Finally, we also show that our model is able to perform zero-shot and value-absent inference. Our trained models using MXT framework are being used to extract attributes for over 12000 PT-attributes in a popular e-commerce store, and have extracted more than 150MM attribute values.

\section{MXT Framework}
\label{sec:mxt}
Given a set of product-types (PTs) $\mathcal{P}=\{p_1,p_2,\ldots,p_m\}$ and  attribute-names $\mathcal{A} = \{a_1,a_2,\ldots,a_n\}$,
we define $\text{MXT}_{\mathcal{P},\mathcal{A}}$ as a multi-PT, multi-attribute, and multi-modal generative model that is trained on PT-attributes from $\left(\mathcal{P},\mathcal{A}\right)$,
and can be used to generate attribute value for any product in the trained PT-attribute set. The 
overall architecture of our model is described in figure~\ref{fig:mmpag_arch}. 
\subsection{Problem Formulation}
We formulate the problem of attribute extraction as the problem of answer generation given a question and a context. Here question is the attribute-name $a\in\mathcal{A}$, and context consists of textual description, product type $p\in\mathcal{P}$ and image 
of the product. All of these are used to extract attribute values. 
The answer generated from the 
model is the attribute value for $a$. 
As shown in figure~\ref{fig:mmpag_arch}, our model architecture mainly 
consists of 3 components: (a) Image-aware Text encoder, (b) Attribute-aware Text-Image Fusion, and (c) Text decoder. 
Below, we describe each component in detail.
\begin{figure}[h]
  \centering
    \includegraphics{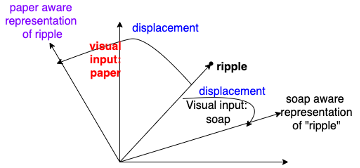}
    \caption{Shift in text embeddings (e.g. "ripple") after applying MAG with visual embeddings}
  \label{fig:pic_vis}
\end{figure}
\subsection{Image-aware Text encoder}
We use T5~\cite{Raffel2020ExploringTL}, which is a transformer~\cite{Vaswani2017AttentionIA} based text only Seq2Seq pretrained language model. 
It includes a bidirectional encoder and a unidirectional (forward only) decoder. In this section, we give an overview of T5's encoder and details 
of its usage for our task. Our text input consists of (i) attribute-name (e.g. "color"), (ii) product-type (e.g. "dress"), and (iii) textual 
description of product. In our QnA format, the question consists of  attribute-name, 
and context consists of concatenation of product-type and textual description of the product. We tokenize both question and context and create a single input sequence of tokens.
This input sequence $x$ is then fed to an embedding and positional 
encoding layer to create input features $T_{emb}\in\mathbb{R}^{N\times d}$, where $N$ is the sequence length and $d$ is the feature dimension. These 
input text embeddings are then fused with Multimodal Adaptation Gate (MAG) as described in Rehman et. al.~\cite{rahman-etal-2020-integrating} to 
generate image aware text embeddings. Due to MAG, the internal representation of words (at any transformer layer) is shifted conditioned on 
visual modalities. This attachment essentially puts words into a different semantic space, which is conditioned on the visual inputs. 
For e.g., the meaning of the word “ripple” changes according to the visual input soap image or paper image. With soap, the meaning 
is “free and clear”, while with paper, the meaning is “wavy pattern” as shown in figure \ref{fig:pic_vis}. This module shifts the meaning
of “ripple” according to visual modality. Since T5 is pretrained model and can understand only text embeddings it is required to fuse the 
visual embeddings ($V_R\in\mathbb{R}^{d}$) with text before feeding it to T5 Encoder rather than feeding the visual embeddings along with text. 
Specifically, in MAG, for each input token $i$ of the sequence, we first learn a gating vector $g_i$ using concatenated embeddings 
of $T_{emb}^i$ and $V_R$: $g_i = RELU(W_g[T_{emb}^i; V_R] + b_g)$. This gating vector highlights the relevant information in visual modality 
conditioned on the input textual vector. We then create an image displacement vector $H_i$ by multiplying $V_R$ with each token's gating vector
$g_i$: $H_i = g_i \cdot (W_H V_R) + b_H$. Finally, we shift the embedding $T_{emb}^i$ by the weighted displacement vector $H_i$ to get the 
multimodal vector $\hat{T}_{emb}^i = T_{emb}^i + \alpha * H_i$. In this equation, $\alpha = min ( \frac{||T_{emb}^i||_2}{||H_i||_2} * \beta, 1)$, 
where $\beta$ is a hyper-parameter whose value is taken as it is from the paper~\cite{rahman-etal-2020-integrating}. This is then passed 
through a layer normalization followed by a dropout layer to get the final fused embedding $F_{MAG}$ from MAG module, where $F_{MAG}^i = \text{dropout}(\text{LN}(\hat{T}_{emb}^i))$. This fused output is then fed to the T5 encoder. The encoder consists of $L$ encoder-layers. It takes $F_{MAG}$ as input gives $T_{enc}$ as output. 
Equation~\ref{eq:t5_encoder} shows the encoding done by $k^{th}$ layer. Here SA is the multi-head self attention layer, Res is the residual 
connection, LN is the layer normalization, and FC is a fully connected layer.
\begin{align}
\label{eq:t5_encoder}
T_{enc}^k = \text{LN}(\text{Res}(\text{FC}(\text{LN}(\text{Res}(\text{SA}(T_{enc}^{k-1}))))))
\end{align}

\subsection{Attribute-aware Text-Image Fusion}
Xception\cite{chollet2017xception} model performs depth-wise (or channel-wise) separable convolutions, i.e., it applies separate 
filters for different color channels. We propose another fusion layer based on the Xception network. The advantage of using this is that it can readily learn the visual features conditioned 
on the attribute type. For example, for the attribute “sleeve type” of a dress, it can identify the channel/color difference between 
sleeves of dress and skin of the person, thus identifying whether sleeve is half or full. We then fuse the text and image embeddings 
using multi-head cross attention. As shown in figure~\ref{fig:mmpag_arch}(b), a product image has several regions of interest, 
for different attributes like "neck style" and "sleeve type". This region specific embeddings are learnt by separable convolutions 
in Xception which is then attended with text embeddings to arrive at attribute aware text embeddings. Now given text 
embedding $T_{enc}\in\mathbb{R}^{N\times d}$ and image embedding $V_X\in\mathbb{R}^{1\times x}$ (from MXT), we create an attribute-aware fused embedding 
$F_A\in\mathbb{R}^{N\times d}$ (having same dimension as of text embedding). This fused embedding is created through a multi-head cross attention module, that applies cross attention between textual and visual embeddings as shown in figure~\ref{fig:mmpag_arch}.
This fusion has an advantage that for an attribute, 
different attention scores can be learned for each object of an image, allowing attending to specific portions of the product image 
conditioned on the attribute name in the question. For example, for the product type "shirt" and attribute "sleeve-type", we may want 
to concentrate only on the portions of the image where sleeves are visible.

\subsection{Text Decoder}
We use T5's unidirectional decoder to output the attribute values. The input to the decoder is the fused embedding vector 
$F_A=<F^1_A, F^2_A,\ldots,F^N_A>$. The decoder iteratively attends to previously generated tokens $y_{<j}$ (via self-attention) and $F_A$ 
(via cross-attention), then predicts the probability of future text tokens $P_{\theta}(y_j | y_{<j} , x, I) = \textrm{Dec}(y_{<j} , F_A)$.
For attribute generation, we fine-tune our model parameters $\theta$ by minimizing the negative log-likelihood of label text $y$ tokens given 
input text $x$ and image $I$: $L_{\theta}^{GEN} = -\sum_{j=1}^{|y|}log P_{\theta}(y_j | y_{<j}, x, I)$.
\section{Experimental Setup \& Results}
\label{sec:experiments}
\textbf{30PT Dataset:} We picked 30 product types (PTs) consisting of total 38 unique attributes from a popular e-commerce store. For each product in the dataset, we have textual information and image. 
The dataset has 569k and 84k products in train and validation data across 30 PTs. 
Our test data consists of products from two product types with a total of 73k products. 
\begin{table*}[]
\centering
  \begin{minipage}[b]{0.6\linewidth}
    \small
    \centering
    \begin{tabular}{ccccccc}
    \toprule
    \multirow{2}{*}{\textbf{PT}} & \multirow{2}{*}{\textbf{\begin{tabular}[c]{@{}c@{}}\#top\\ attributes\end{tabular}}} & \multirow{2}{*}{\textbf{\begin{tabular}[c]{@{}c@{}}CMA-\\ CLIP\end{tabular}}} & \multicolumn{4}{c}{\textbf{MXT}}                                                                                                                                                                                                              \\ \cmidrule{4-7} 
                                 &                                                                                      &                                                                               & \multicolumn{1}{c}{\textbf{Multi-PT}}     & \multicolumn{1}{c}{\textbf{Single PT}} & \multicolumn{1}{c}{\textbf{\begin{tabular}[c]{@{}c@{}}Without-\\ Xception\end{tabular}}} & \textbf{\begin{tabular}[c]{@{}c@{}}Without-\\ MAG\end{tabular}} \\ \midrule
    \multirow{3}{*}{A}           & K=5                                                                                  & +6.16\%                                                                       & \multicolumn{1}{c}{\textbf{+22.33\%}} & \multicolumn{1}{c}{+19.58\%}           & \multicolumn{1}{c}{+21.82\%}                                                             & +20.93\%                                                        \\ \cmidrule{2-7} 
                                 & K=10                                                                                 & +6.70\%                                                                       & \multicolumn{1}{c}{\textbf{+16.89\%}} & \multicolumn{1}{c}{+15.50\%}           & \multicolumn{1}{c}{+15.60\%}                                                             & +15.19\%                                                        \\ \cmidrule{2-7} 
                                 & K=15                                                                                 & +1.81\%                                                                       & \multicolumn{1}{c}{\textbf{+13.23\%}} & \multicolumn{1}{c}{+11.64\%}           & \multicolumn{1}{c}{+10.67\%}                                                             & +10.45\%                                                        \\ \midrule
    \multirow{3}{*}{B}           & K=5                                                                                  & +8.34\%                                                                       & \multicolumn{1}{c}{\textbf{+16.63\%}} & \multicolumn{1}{c}{+12.86\%}           & \multicolumn{1}{c}{+13.94\%}                                                             & +13.58\%                                                        \\ \cmidrule{2-7} 
                                 & K=10                                                                                 & +18.46\%                                                                      & \multicolumn{1}{c}{\textbf{+24.98\%}} & \multicolumn{1}{c}{+22.46\%}           & \multicolumn{1}{c}{+22.81\%}                                                             & +22.55\%                                                        \\ \cmidrule{2-7} 
                                 & K=15                                                                                 & +11.72\%                                                                      & \multicolumn{1}{c}{\textbf{+18.51\%}} & \multicolumn{1}{c}{+15.50\%}           & \multicolumn{1}{c}{+16.28\%}                                                             & +15.68\%                                                        \\ \bottomrule
    \end{tabular}
    \end{minipage}\hfill
  \begin{minipage}[b]{0.4\linewidth}
    \centering
    \begin{tabular}{cc}
    \toprule
    \textbf{PT} & \textbf{MXT}      \\ \midrule
    A           & \textbf{+15.56\%} \\ \midrule
    B           & -1.47\%           \\ \midrule
    C           & -7.89\%           \\ \midrule
    D           & \textbf{+9.98\%}  \\ \midrule
    E           & \textbf{+13.23\%} \\ \bottomrule
    \end{tabular}
    \end{minipage}
  \caption{Left: Improvement in Recall@90P\% of CMA-CLIP and MXT (with different ablation studies) over NER-MQMRC on 30PT datasetE-commerce5PT dataset. Right: Improvement in F1-score of MXT over NER-MQMRC on E-commerce5PT dataset}
  \label{fig:twotables}
\end{table*}

We evaluated MXT against two state of the art methods on attribute extraction: (1) \textbf{CMA-CLIP: } A multi-task classifier that uses
CLIP~\cite{Radford2021LearningTV} for learning multi-modal embeddings of products followed by using two types of cross-modality attentions: 
(a) sequence-wise attention to capture relation between individual text tokens and image features, and (b) modality-wise attention to capture 
weightage of text and image features relative to each downstream task, (2) \textbf{NER-MQMRC: } This framework~\cite{Shrimal2022NERMQMRCFN} poses Named Entity Recognition (NER) problem as Multi Question 
Machine Reading Comprehension (MQMRC) task. This is the state of the art model for the text-based attribute extraction task. In 
this model, given the text description of a product (\emph{context}), they give attribute names as \emph{multiple questions} to 
their BERT based MRC architecture, which finds span of each attribute value \emph{answer} from the context.

Left table in the figure~\ref{fig:twotables} compares the recall@90P\% of the three models. 
We show the performance on top-5, top-10 and top-15 attributes (by number of products in which they are present.
We can see that MXT outperforms MQMRC and CMA-CLIP on both product types.

\textbf{E-commerce5PT:} This is a benchmark dataset from NER-MQMRC paper~\cite{Shrimal2022NERMQMRCFN}. We take a subset of this dataset (removing numerical attributes) consisting of 22 product-attributes across 5 product types. This is a benchmark dataset for NER based models since all attribute values are present in the text in this dataset. The dataset has 273,345 and 4,259 products in train and test data respectively. We compare average F1 scores (averaged across attributes for each product type) of MXT model with NER-MQMRC on this dataset where our model outperforms NER-MQMRC on 16/22 attributes. 
Right table in the figure~\ref{fig:twotables} shows the average F1-scores (across attributes in each product type) of MXT and NER-MQMRC models.

\subsection{Ablation Study}
We show three ablation studies on 30PT dataset that justify our choices in the MXT architecture. Left table in the figure~\ref{fig:twotables} shows the results of these studies. \textbf{(a) Scalability:} We show that our proposed framework is highly scalable. For that, we compute Recall@90P\% of the MXT model trained on individual PTs. The results show that (i) our model leverages cross-PT information during training, (ii) we don't need to train 
separate model for each PT, which makes model monitoring and refreshing easier in the production, \textbf{(b) Xception network:} We show that Xception network helps concentrating on certain attribute features. For this, we removed the Xception network from our architecture and trained and evaluated the model, 
\textbf{(c) MAG:} We replaced MAG with simple concatenation of text and image embeddings in MXT. We can see in the table that each of our ablation model under-performs the MXT model trained on 30PTs, thus justifying our design choices.


\subsection{Zero-shot Inference and Value-absent Inference}
\label{subsec:zero_shot}
Most existing methods for attribute extraction face two challenges: \textbf{(i) Zero-shot inference:} All the predictive models (classification-based models) 
can predict attribute values only from a predefined set of values
that are seen in the training data. They are unable to do zero-shot inference i.e. they can't predict an attribute value if it is not seen in the
training data, \textbf{(ii) Value-absent inference:} All NER-based models can extract values only
which are mentioned in the text data i.e. if an attribute value is absent in the input text, they can't extract that value. Our generative model 
solves both of these challenges. 
For example, in the E-commerce5PT dataset, there are a total of 8289 product-attribute pairs in the test data, out of which 
970 product-attribute pairs were not seen in the training data, from which our model correctly generated 124 product-attribute pairs. For example, given
a product of product-type \emph{"dress"} with title \emph{"Tahari ASL Women's Sleeveless Ruched Neck Dress with Hi Lo Skirt"}, our model
generated the value \emph{"Ruched Neck"} for the attribute \emph{"neck style"}. Here the value \emph{"Ruched Neck"} was absent from the training data.
Similarly, for the \emph{"dress"} product shown in figure~\ref{fig:intro-fig}   
, our model generated the value \emph{"mini"} for the attribute \emph{"item length"} (by inferring it from the image)
even when this value is not mentioned in the product text(thus solving the second challenge).

\subsection{Training \& Inference Details}
We conducted training for each model over a span of 20 epochs, employing a batch size of 4.
The training process was performed using distributed multi-GPU training across 8 V-100 Nvidia GPUs, each equipped with 16GB of memory. 
For text encoder and decoder, we finetune the pretrained \texttt{t5-base}~\footnote{The \texttt{t5-base} checkpoint is available at \url{https://huggingface.co/transformers/model_doc/t5.html}} checkpoint.
We obtained ResNet-based image embeddings using a pretrained ResNet-152, specifically with one embedding assigned to each image. ~\footnote{\url{https://download.pytorch.org/models/resnet152-b121ed2d.pth}}. During training, we employed the Adam optimizer with learning rate of $5e^{-5}$ and warmup ratio of 0.1. We chose the checkpoint having best validation loss. For inference, we used greedy search to generate attribute values.
\section{Deployment}
\label{sec:deployment}
In a popular e-commerce store, we have deployed MXT for 6 English speaking markets covering >10K PT-attributes and have extracted >150MM attribute values.\\
\textbf{Design Choices:} In popular e-commerce stores, usually there are more than 100K PT-attributes across various markets. Earlier models like NER-MQMRC or CMA-CLIP could be trained only for few 100s of PT-attributes. NER-MQMRC~\cite{shrimal-etal-2022-ner} architecture only allowed one product type in one model training, while CMA-CLIP couldn't scale beyond few 100s of PT-attribute pairs due to network explosion (as they had to create an output layer for each of the different attribute value). This had serious issues of monitoring, refreshing and maintaining the quality of models. Our prompt-based approach in MXT allows us to train a single model checkpoint for any number of PT-attribute pairs.\\
\textbf{Practical Challenges:} We faced several challenges during building and deploying the model. One of the biggest challenge was lack of normalized attribute values. Since we were relying on the distantly supervised training data from the catalog, there were multiple junk values. Normalizing these values is challenging without the support of annotations. To overcome this problem, we used some heuristic matches to merge similar values. We also trimmed the tail attribute values to remove the junk values further. The second major challenge was to evaluate the model and find the threshold for every PAC to achieve the desired precision. Since we had >10K PT-attributes, even if we annotate 300 samples per PT-attribute, it leads to 3MM annotations, which is not feasible. For that, we evaluated the model automatically using the catalog data. Since the catalog data can be noisy, we checked other things like whether the predicted value is present in text, whether the attribute should allow zero-shot prediction etc. Based on these checks, we decided the required precision accordingly.  
\section{Conclusion \& Future Work}
\label{sec:conclusion}
In this paper, we presented MXT, a large scale multi-modal product attribute generation system to extract product attributes from the products listed in eCommerce stores. Our model infers the attribute values using both textual and visual information present on the product pages. We introduced a novel architecture comprising a T5 based encoder and decoder along with two fusion layers to fuse text and image embeddings. We showed our model can beat the existing state of the art extractive as well as predictive models on the benchmark datasets. Our model is scalable to multiple product types and countries by just specifying them in the input text prompt. We further showed that our model is able to perform zero-shot inference, as well as it can generate attribute values not present in the text.
There are several future directions to explore which can further improve the performance of our model. First, we would like to create an ensemble of NER-based and generative models so that we can leverage the power of extraction based models which work very well for numerical attributes (e.g. size, length etc.). Second, our current approach does not use relational information among the products. Since similar products can have common attribute values, we can use graph based approaches to capture that relational information. Specifically, we can approach the attribute extraction problem through either link prediction or node classification. In the former method, we aim to predict missing links between products and their attributes. Alternatively, the latter approach involves using similarity between product features, including text, images, and co-viewing information, to determine graph edges for classification of product nodes.
\section{Limitations}
\label{sec:limitations}
In this section, we discuss some of the limitations of our current model architecture: (1) \textbf{Non-English locales:} Currently in our experiments, we have trained and evaluated models only on English datasets. Building models on non-English locales is the direction for future work, (2) \textbf{Use of pre-trained tokenizer: } The T5's tokenizer in our models has been pre-trained on open-domain datasets, and its vocabulary misses out on e-commerce specific terms. For example, the current tokenizer of T5 tokenizes the phrase “skater midi dress” as [“sk”, “a”, “ter”, “mid”, “I”, “dress”]. Here, the meaning of words “skater” and “midi” is not captured in the tokenized text. We believe that we can overcome this limitation by pre-training T5 on e-commerce data which would help tokenizer understanding and tokenizing the e-commerce specific terms more correctly.

\section{Acknowledgements}
We thank all the anonymous reviewers for providing their valuable comments that helped us
improve the quality of our paper. We also thank our colleagues in the science, product, and engineering teams at Amazon for their valuable
inputs.

\bibliography{anthology,custom,reference}

\begin{thebibliography}{15}
\expandafter\ifx\csname natexlab\endcsname\relax\def\natexlab#1{#1}\fi

\bibitem[{Chollet(2017)}]{chollet2017xception}
Fran{\c{c}}ois Chollet. 2017.
\newblock Xception: Deep learning with depthwise separable convolutions.
\newblock In \emph{Proceedings of the IEEE conference on computer vision and
  pattern recognition}, pages 1251--1258.

\bibitem[{He et~al.(2016)He, Zhang, Ren, and Sun}]{He2016DeepRL}
Kaiming He, X.~Zhang, Shaoqing Ren, and Jian Sun. 2016.
\newblock Deep residual learning for image recognition supplementary materials.
\newblock In \emph{IEEE conference on computer vision and pattern recognition},
  pages 770--778.

\bibitem[{Lin et~al.(2021)Lin, He, Feng, Zalmout, Liang, Xiong, and
  Dong}]{Lin2021PAMUP}
Rongmei Lin, Xiang He, Jie Feng, Nasser Zalmout, Yan Liang, Li~Xiong, and Xin
  Dong. 2021.
\newblock Pam: Understanding product images in cross product category attribute
  extraction.
\newblock \emph{Proceedings of the 27th ACM SIGKDD Conference on Knowledge
  Discovery \& Data Mining}.

\bibitem[{Liu et~al.(2021)Liu, Xu, Fu, Liu, Xie, Wang, Wang, and
  Sun}]{Liu2021CMACLIPCA}
Huidong Liu, Shaoyuan Xu, Jinmiao Fu, Yang Liu, Ning Xie, Chien Wang, Bryan
  Wang, and Yi~Sun. 2021.
\newblock Cma-clip: Cross-modality attention clip for image-text
  classification.
\newblock \emph{ArXiv}, abs/2112.03562.

\bibitem[{Mehta et~al.(2021)Mehta, Oprea, and
  Rasiwasia}]{Mehta2021LATEXNumericLA}
Kartik Mehta, Ioana Oprea, and Nikhil Rasiwasia. 2021.
\newblock Latex-numeric: Language agnostic text attribute extraction for
  numeric attributes.
\newblock In \emph{NAACL}.

\bibitem[{Radford et~al.(2021)Radford, Kim, Hallacy, Ramesh, Goh, Agarwal,
  Sastry, Askell, Mishkin, Clark, Krueger, and
  Sutskever}]{Radford2021LearningTV}
Alec Radford, Jong~Wook Kim, Chris Hallacy, Aditya Ramesh, Gabriel Goh,
  Sandhini Agarwal, Girish Sastry, Amanda Askell, Pamela Mishkin, Jack Clark,
  Gretchen Krueger, and Ilya Sutskever. 2021.
\newblock Learning transferable visual models from natural language
  supervision.
\newblock In \emph{ICML}.

\bibitem[{Raffel et~al.(2020)Raffel, Shazeer, Roberts, Lee, Narang, Matena,
  Zhou, Li, and Liu}]{Raffel2020ExploringTL}
Colin Raffel, Noam~M. Shazeer, Adam Roberts, Katherine Lee, Sharan Narang,
  Michael Matena, Yanqi Zhou, Wei Li, and Peter~J. Liu. 2020.
\newblock Exploring the limits of transfer learning with a unified text-to-text
  transformer.
\newblock \emph{ArXiv}, abs/1910.10683.

\bibitem[{Rahman et~al.(2020{\natexlab{a}})Rahman, Hasan, Lee, Zadeh, Mao,
  Morency, and Hoque}]{Rahman2020IntegratingMI}
Wasifur Rahman, M.~Hasan, Sangwu Lee, Amir Zadeh, Chengfeng Mao, Louis-Philippe
  Morency, and Ehsan Hoque. 2020{\natexlab{a}}.
\newblock Integrating multimodal information in large pretrained transformers.
\newblock \emph{Proceedings of the conference. Association for Computational
  Linguistics. Meeting}, 2020:2359--2369.

\bibitem[{Rahman et~al.(2020{\natexlab{b}})Rahman, Hasan, Lee, Bagher~Zadeh,
  Mao, Morency, and Hoque}]{rahman-etal-2020-integrating}
Wasifur Rahman, Md~Kamrul Hasan, Sangwu Lee, AmirAli Bagher~Zadeh, Chengfeng
  Mao, Louis-Philippe Morency, and Ehsan Hoque. 2020{\natexlab{b}}.
\newblock \href {https://doi.org/10.18653/v1/2020.acl-main.214} {Integrating
  multimodal information in large pretrained transformers}.
\newblock In \emph{Proceedings of the 58th Annual Meeting of the Association
  for Computational Linguistics}, pages 2359--2369, Online. Association for
  Computational Linguistics.

\bibitem[{Roy et~al.(2021)Roy, Goyal, and Pandey}]{roy-etal-2021-attribute}
Kalyani Roy, Pawan Goyal, and Manish Pandey. 2021.
\newblock \href {https://doi.org/10.18653/v1/2021.ecnlp-1.2} {Attribute value
  generation from product title using language models}.
\newblock In \emph{Proceedings of the 4th Workshop on e-Commerce and NLP},
  pages 13--17, Online. Association for Computational Linguistics.

\bibitem[{Shrimal et~al.(2022{\natexlab{a}})Shrimal, Jain, Mehta, and
  Yenigalla}]{shrimal-etal-2022-ner}
Anubhav Shrimal, Avi Jain, Kartik Mehta, and Promod Yenigalla.
  2022{\natexlab{a}}.
\newblock \href {https://doi.org/10.18653/v1/2022.naacl-industry.26}
  {{NER-MQMRC}: {F}ormulating named entity recognition as multi question
  machine reading comprehension}.
\newblock In \emph{Proceedings of the 2022 Conference of the North American
  Chapter of the Association for Computational Linguistics: Human Language
  Technologies: Industry Track}, pages 230--238, Hybrid: Seattle, Washington +
  Online. Association for Computational Linguistics.

\bibitem[{Shrimal et~al.(2022{\natexlab{b}})Shrimal, Jain, Mehta, and
  Yenigalla}]{Shrimal2022NERMQMRCFN}
Anubhav Shrimal, Avi~Rajesh Jain, Kartik Mehta, and Promod Yenigalla.
  2022{\natexlab{b}}.
\newblock Ner-mqmrc: Formulating named entity recognition as multi question
  machine reading compmrehension.
\newblock \emph{ArXiv}, abs/2205.05904.

\bibitem[{Vaswani et~al.(2017)Vaswani, Shazeer, Parmar, Uszkoreit, Jones,
  Gomez, Kaiser, and Polosukhin}]{Vaswani2017AttentionIA}
Ashish Vaswani, Noam~M. Shazeer, Niki Parmar, Jakob Uszkoreit, Llion Jones,
  Aidan~N. Gomez, Lukasz Kaiser, and Illia Polosukhin. 2017.
\newblock Attention is all you need.
\newblock \emph{ArXiv}, abs/1706.03762.

\bibitem[{Yu et~al.(2021)Yu, Dai, Liu, and Fung}]{Yu2021VisionGG}
Tiezheng Yu, Wenliang Dai, Zihan Liu, and Pascale Fung. 2021.
\newblock Vision guided generative pre-trained language models for multimodal
  abstractive summarization.
\newblock \emph{ArXiv}, abs/2109.02401.

\bibitem[{Zheng et~al.(2018)Zheng, Mukherjee, Dong, and
  Li}]{Zheng2018OpenTagOA}
Guineng Zheng, Subhabrata Mukherjee, Xin Dong, and Feifei Li. 2018.
\newblock Opentag: Open attribute value extraction from product profiles.
\newblock \emph{Proceedings of the 24th ACM SIGKDD International Conference on
  Knowledge Discovery \& Data Mining}.

\end{thebibliography}
\bibliographystyle{acl_natbib}
\clearpage
\end{document}